\documentclass[]{sn-jnl}% APA Reference Style 

\usepackage{graphicx}%
\usepackage{multirow}%
\usepackage{amsmath,amssymb,amsfonts}%
\usepackage{amsthm}%
\usepackage{mathrsfs}%
\usepackage{multicol}
\usepackage{multirow}
\usepackage[title]{appendix}%
\usepackage{xcolor}%
\usepackage{textcomp}%
\usepackage{manyfoot}%
\usepackage{booktabs}%
\usepackage{algorithm}%
\usepackage{algorithmicx}%
\usepackage{algpseudocode}%
\usepackage{listings}%
\usepackage{lineno} % Add line numbers package

\usepackage{geometry}
\usepackage{subfigure}
\usepackage{amsmath} % for multline
\usepackage{tikz}
\usepackage{caption}
\usepackage{adjustbox}
\usetikzlibrary{shapes.geometric, arrows}
\tikzstyle{startstop} = [rectangle, rounded corners, 
minimum width=3cm, 
minimum height=1cm,
text centered, 
draw=black, 
fill=red!30]

\tikzstyle{io} = [trapezium, 
trapezium stretches=true, % A later addition
trapezium left angle=70, 
trapezium right angle=110, 
minimum width=3cm, 
minimum height=1cm, text centered, 
draw=black, fill=blue!30]

\tikzstyle{process} = [rectangle, 
minimum width=3cm, 
minimum height=1cm, 
text centered, 
text width=3cm, 
draw=black, 
fill=orange!30]

\tikzstyle{decision} = [diamond, 
minimum width=3cm, 
minimum height=2cm,
text centered, 
draw=black, 
fill=green!30]
\tikzstyle{arrow} = [thick,->,>=stealth]
\usepackage[style=apa, backend=biber]{biblatex}
\theoremstyle{thmstyleone}%
\theoremstyle{thmstyletwo}%

\theoremstyle{thmstylethree}%

\begin{document}

\nolinenumbers

\title[Article Title]{Adaptive grid-based decomposition for UAV-based coverage path planning in maritime search and rescue}


\author*{\fnm{Sina} \sur{Kazemdehbashi}

\text{hm3369@wayne.edu}}


\abstract{

Unmanned aerial vehicles (UAVs) are increasingly utilized in search and rescue (SAR) operations to enhance efficiency by enabling rescue teams to cover large search areas in a shorter time. Reducing coverage time directly increases the likelihood of finding the target quickly, thereby improving the chances of a successful SAR operation. In this context, UAVs require path planning to determine the optimal flight path that fully covers the search area in the least amount of time. A common approach involves decomposing the search area into a grid, where the UAV must visit all cells to achieve complete coverage. In this paper, we propose an Adaptive Grid-based Decomposition (AGD) algorithm that efficiently partitions polygonal search areas into grids with fewer cells. Additionally, we utilize a Mixed-Integer Programming (MIP) model, compatible with the AGD algorithm, to determine a flight path that ensures complete cell coverage while minimizing overall coverage time. Experimental results highlight the AGD algorithm’s efficiency in reducing coverage time (by up to $ 20 \%$) across various scenarios.
}




\maketitle

\section{Introduction}\label{sec:intro}

Today, Search and Rescue (SAR) teams are increasingly leveraging advanced technologies such as artificial intelligence and Unmanned Aerial Vehicles (UAVs) to enhance the efficiency of their operations \parencite{martinez2021search}. In this context, UAVs, with their high flight speed and ability to scan areas at night or in low-light conditions, can address one of the challenges in SAR operations: monitoring large or hard-to-reach search areas. In ground SAR operations, additional methods such as employing dogs and volunteers can be used alongside UAVs to expedite target searching. However, in maritime SAR operations, fewer options are available, making UAVs particularly important for enhancing operational efficiency. In this regard, one of the main questions is how UAVs should fly to cover the search area in the shortest possible time, a challenge addressed in the literature under the \emph{Coverage Path Planning (CPP)} problem. Various objective functions were considered in CPP, including the number of turning maneuvers \parencite{maza2007multiple}, path length (\cite{Bouzid2017}), flight time (\cite{forsmo2013optimal}), energy consumption (\cite{di2016coverage}), and total coverage time (\cite{kazemdehbashi2025algorithm}). Additionally, two main types of decomposition are used in the CPP problem: exact cell decomposition and grid-based decomposition. In exact cell decomposition, the search area is divided into smaller sub-areas, whereas in grid-based decomposition, the area is represented as a grid, and each grid's cell must be covered to achieve full coverage. In this paper, we propose an Adaptive Grid-based Decomposition (AGD) algorithm to reduce the number of cells in the grid required to cover the primary search area. We also review a Mixed Integer Programming (MIP) model that employs the AGD algorithm to generate paths for a single UAV.

The organization of the paper is as follows: Section \ref{sec: literature-review} reviews the relevant literature. Section \ref{sec:solution}
 first defines the problem, then explains the AGD algorithm, and finally reviews an MIP model adapted to the AGD algorithm for solving the CPP with a single UAV. In Section \ref{sec:experimental}, we conduct experiments on various cases to demonstrate the efficiency of the proposed AGD algorithm. Finally, Section \ref{sec:conclusion} summarizes the findings and suggests potential developments for future work.

\section{Literature Review}\label{sec: literature-review}

Today, UAVs are increasingly employed in various fields such as delivery, search and rescue (SAR), and agriculture. In SAR operations, UAV path planning aimed at minimizing the survey time of the search area can be addressed as a CPP problem. The CPP problem can be classified based on the decomposition type, which refers to how the search area is divided into manageable sub-areas (\cite{cabreira2019survey}). Since different decomposition types require different mathematical approaches to address the CPP problem, we categorize CPP methods into two main groups based on their decomposition methods: exact cellular decomposition and approximate (grid-based) decomposition of the search area.

In the exact cell decomposition method, the main area is divided into sub-areas, called cells. \textcite{Latombe1991} proposed a trapezoidal decomposition in which the area is divided into trapezoidal cells. Each cell can be covered by simple back-and-forth motions. 
\textcite{Choset2000} introduced the boustrophedon decomposition, an enhanced approach aimed at combining trapezoidal cells into larger, non-convex cells. This action provides opportunities to reduce the path length during the robots' back-and-forth movements to cover the cells.
\textcite{Coombes2018} introduced a novel approach for planning coverage paths in fixed-wing UAV aerial surveys. They took windy conditions into account and demonstrated that flying perpendicular to the wind direction offers a flight time advantage compared to flying parallel to it. Furthermore, they applied dynamic programming to identify the time-optimal convex decomposition within a polygon.
\textcite{bahnemann2021revisiting} extended on boustrophedon coverage planning by exploring various sweep directions within each cell to determine the most efficient sweep path. They also used the Equality Generalized Traveling Salesman Problem (E-GTSP) to model and identify the shortest possible path in the adjacency graph.

In the following paragraph, we review studies that employed approximate (grid-based) decomposition methods.

As mentioned before, in this approach, the main search area is divided into a grid, and by covering all the grid's cells, the entire search area will be fully surveyed. \textcite{Valente2013} developed a path planning tool that transforms irregular areas into grid graphs, generating near-optimal UAV trajectories by reducing the number of turns in their paths. \textcite{Cho2021} proposed a grid-based decomposition to convert an area into a graph. They formulated the problem as a MIP model where minimizing the completion time of the coverage path is the objective function. Also, they presented a randomized search heuristic (RSH) algorithm to reduce the computation time for large-scale instances while maintaining a small optimality gap. Their study focused on utilizing heterogeneous UAVs in polygonal areas for maritime search and rescue missions. \textcite{ahmed2023energy} addressed the CPP problem for multiple UAVs by developing two methods: a greedy algorithm and Simulated Annealing (SA). They used a MIP model to formulate the problem for minimizing energy consumption. They showed that for large-scale cases SA had better performance to reduce the overall energy consumption, but in small-scale cases, CPLEX was the best. 
\textcite{kazemdehbashi2025algorithm} proposed an exact bound algorithm to find a set of paths for a fleet of UAVs in a rectangular search area under windy conditions. Also, they presented a formula to calculate the lower bound of the coverage time based on the search area dimensions and number of UAVs.

The contributions of our research are summarized as follows. First, we propose the Adaptive Grid-based Decomposition (AGD) algorithm to decompose the search area into a grid with fewer cells. Second, we review a MIP model adapted to the AGD algorithm to address the CPP problem for a single UAV and conduct numerical experiments to evaluate the algorithm's performance.

\section{Problem Description and Solutions} \label{sec:solution}

The primary objective of SAR operations is to locate the target as quickly as possible. In maritime environments, time is even more critical than on land. If the target loses energy and submerges, survival becomes unlikely, and aerial images can no longer detect it. UAVs play a crucial role in rapidly capturing aerial images over large areas, assisting SAR teams in locating targets more efficiently. To achieve this, the search area must first be decomposed into a grid (or a graph where nodes represent the centers of the cells), and then a mathematical model should determine the order in which cells (or graph nodes) are covered, ensuring full coverage of the search area. Based on this approach, we first propose a grid-based decomposition method, followed by the use of a MIP model to generate a coverage path for a single UAV.
\begin{figure}
    \centering
    \subfigure(a){\includegraphics[width=0.38\textwidth]{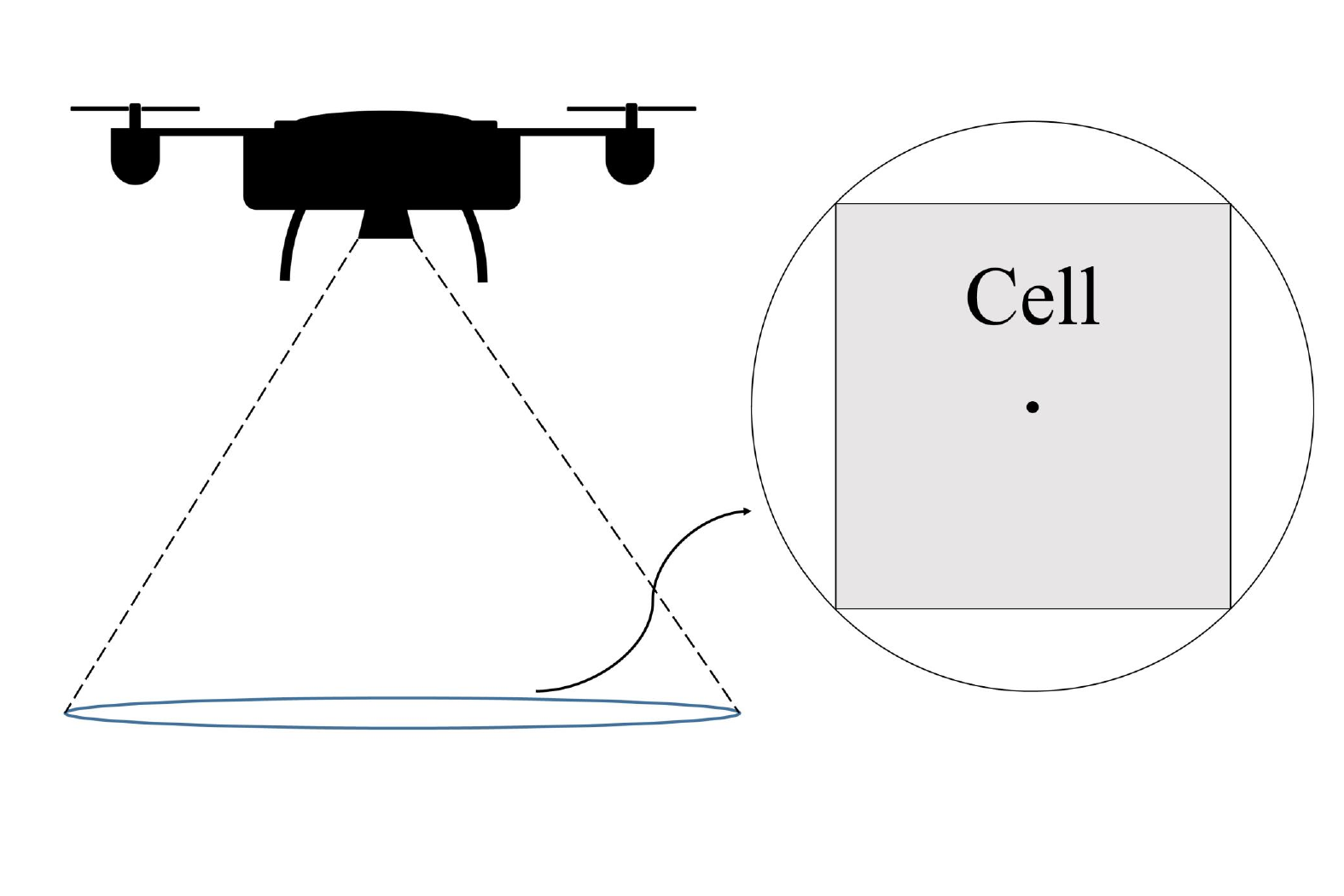}}
    \hspace{0.4 cm}
    \subfigure(b){\includegraphics[width=0.5\textwidth]{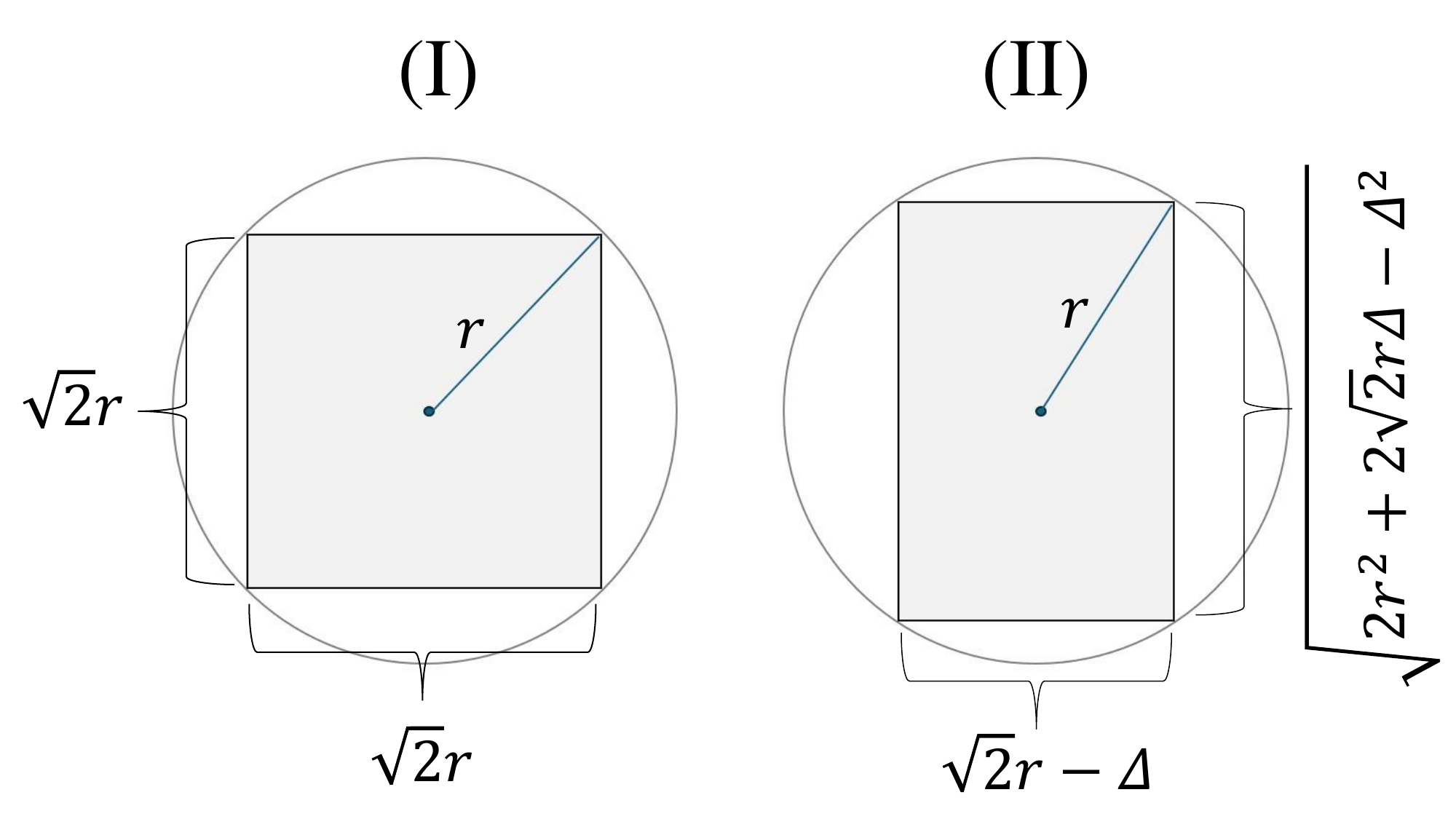}} 
    \caption{(a) The UAV's camera footprint; (b) By considering \( r \) as the radius of the camera's footprint, (I) shows the grid cell dimensions, and (II) shows the grid cell dimensions after a \( \Delta \) adjustment to one edge (subtracting \( \Delta \) from one edge).}
    \label{fig:footprint&grid}
\end{figure}

\subsection{Grid-based decomposition}
We propose the Adaptive Grid-based Decomposition (AGD) algorithm to partition the search area into a grid. Each cell of the grid is defined by its center coordinates and must fit within the UAV's camera footprint (see Fig. \ref{fig:footprint&grid}a). In the AGD algorithm, cell dimensions can be adjusted, provided that the modified cells remain within the boundaries of the UAV's camera footprint (see Fig. \ref{fig:footprint&grid}b). In simple terms, the primary goal of the AGD algorithm is to minimize the number of cells required to cover the search area by allowing flexibility in cell dimensions. We use notations in Table \ref{tab:AGD notation} to explain the AGD algorithm. Before initiating the AGD algorithm, for simplicity, we first transform all the polygon’s vertices so that the longest edge lies on the X-axis, and all vertices are positioned in the first quadrant of the XY-coordinate system (see Example 1 in the Appendix).

In the first iteration, the AGD algorithm creates the initial decomposition channel. This process begins by establishing the bottom line of the decomposition channel along the longest edge (i.e., $y_b = 0$). Subsequently, the top line of the decomposition channel, denoted as \( y_t \), is set at \(\sqrt{2}r \), corresponding to the size of the cell edge ($r$ is the radius of the UAV's camera footprint). Next, the AGD algorithm generates a set of points, referred to as the \( P \) set, which includes: (1) points at the intersection of \( y_b \) and the polygon's edges, (2) points located at the intersection of \( y_t \) and the polygon's edges, and (3) the polygon's vertices with Y-coordinates within the interval \( (y_b, y_t) \). The algorithm then calculates the maximum length of the decomposition channel (\( l \)) using the formula \( l = x_{\text{max}} - x_{\text{min}} \), where \( x_{\text{min}} \) and \( x_{\text{max}} \) are the minimum and maximum X-coordinates among the points in the \( P \) set, respectively. Then, \( n \) is calculated as \( \lceil \frac{l}{\sqrt{2}r} \rceil \), representing the number of cells required to cover the decomposition channel before adjusting the cell's edge size. If these \( n \) cells cover an area larger than the decomposition channel, the excess length ($e$) can be calculated as \( \sqrt{2}rn - l \). This allows us to adjust each cell's edge length by \(\Delta=\frac{e}{n} \) while still covering the decomposition channel. Adjusting one edge of a cell by a value \( \Delta \) changes the size of the other edge to \( \sqrt{2r^2 + 2\sqrt{2}r\Delta - \Delta^2} \) (see Fig. \ref{fig:footprint&grid}b), resulting in a taller cell within the UAV's camera footprint. Next, the AGD algorithm updates $y_t$ to $y_t^{adj}=y_b+\sqrt{2r^2+2\sqrt{2}r\Delta-\Delta^2}$. In the final step, the AGD fills the decomposition channel with \( n \) cells, each having dimensions of \( \sqrt{2}r - \Delta \) and \( \sqrt{2r^2 + 2\sqrt{2}r\Delta - \Delta^2} \) (Fig. \ref{fig:footprint&grid}b), and saves their center coordinates. Before proceeding to the next iteration, the AGD algorithm updates $y_b$ to  $y_t^{\text{adj}}$. If \( y_b < y_{\text{max}} \) (where \( y_{\text{max}} \) is the maximum Y-coordinate among the polygon's vertices), it moves to the next iteration by setting the new \( y_t \) to \( y_b + \sqrt{2}r \) and follows the same pattern as in the previous iteration. After the final iteration, all polygon vertices and saved center coordinates are reverse transformed to restore their original coordinates (see 
 Example 1 in the Appendix).
\begin{figure}
    \centering
    \includegraphics[width=0.86\linewidth]{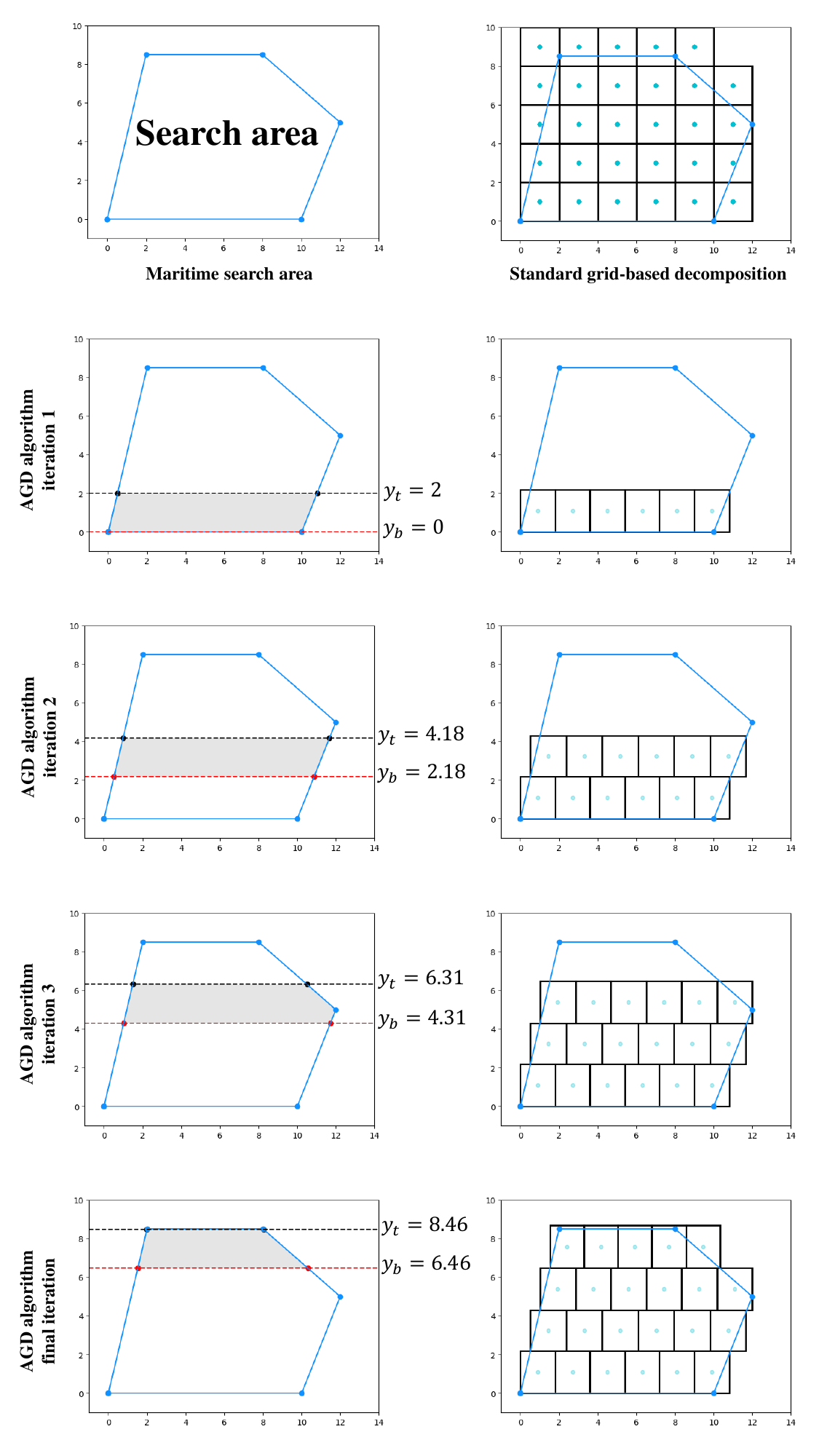}
    \caption{An illustrative example demonstrating the step-by-step implementation of the AGD algorithm. The polygon vertices are $\{(0, 0), (10, 0), (12, 5), (8, 8.5), (2, 8.5)\}$, and the radius of the UAV's camera footprint ($r$) is $\sqrt{2}$. In each iteration, the gray area represents the decomposition channel between \(y_b\) and \(y_t\). Note that in the standard grid-based decomposition of the search area all cells' edge size is $\sqrt{2}r=2$ (see Fig. \ref{fig:footprint&grid}b case (I)).}
    \label{fig:AGD-algorithm}
\end{figure}

We use a simple example to explain the AGD algorithm step by step. Consider a polygon with vertices ordered counterclockwise: \(\{(0,0), (10,0), (12,5), (8,8.5), (2,8.5)\}\) (see Fig. \ref{fig:AGD-algorithm}). Suppose the radius of the UAV's camera footprint (\(r\)) is \(\sqrt{2}\). In the first iteration, the AGD algorithm sets \(y_b = 0\) and \(y_t = (\sqrt{2})^2 = 2\).
The set of points at the intersection of \(y_b\) and the polygon's edges (denoted as \(P_1\)) is \(\{(0,0), (10,0)\}\), while the set of points at the intersection of \(y_t\) and the polygon's edges (denoted as \(P_2\)) is \(\{(0.47, 2), (10.8, 2)\}\). The set of the polygon's vertices with Y-coordinates within the interval \((y_b, y_t)\) (denoted as \(P_3\)) is empty, i.e., \(P_3 = \emptyset\). Thus, the set \(P = P_1 \cup P_2 \cup P_3\) is \(\{(0,0), (10,0), (0.47,2), (10.8,2)\}\). From the set \(P\), we determine \(x_{\text{min}} = 0\) and \(x_{\text{max}} = 10.8\). Therefore, the maximum length of the decomposition channel is calculated as \(l = 10.8 - 0 = 10.8\).
The number of cells in the decomposition channel is \(n = \lceil \frac{10.8}{2} \rceil = 6\). The excess length is \(e = 2(6) - 10.8 = 1.2\), so \(\Delta = \frac{1.2}{6} = 0.2\), which represents the adjustment value for the cell's edge. This adjustment changes the other dimension of the cell to \(2.181\). The AGD algorithm now uses 6 cells, with dimensions \(1.8\) and \(2.181\), to cover the decomposition channel. After adjusting \(y_t\) to \(y_t^{\text{adj}} = 2.181\), the AGD stores the centers of the cells as \(DP = \{(0.9, 1.09), (2.7, 1.09), (4.5, 1.09), (6.3, 1.09), (8.1, 1.09), (9.9, 1.09)\}\).
In the second iteration, \(y_b\) is updated to \(2.181\) (which is \(y_t^{\text{adj}}\) from the previous iteration). Since \(y_b < y_{\text{max}}\) (i.e., \(2.181 < 8.5\)), the AGD algorithm proceeds with the second iteration using \(y_b = 2.181\) and \(y_t = 4.181\).
In this iteration, to find the set \(P\), we have \(P_1 = \{(0.513, 2.18), (10.872, 2.181)\}\), \(P_2 = \{(0.983, 4.181), (11.672, 4.181)\}\), and \(P_3 = \emptyset\). Therefore, \(P = \{(0.513, 2.181), (10.872, 2.181), (11.672, 4.181), (0.983, 4.181)\}\). By considering the points in the set \(P\), we find that \(x_{\text{min}} = 0.513\) and \(x_{\text{max}} = 11.672\). Next, we calculate \(l = 11.672 - 0.513 = 11.159\), \(n = \lceil \frac{11.159}{2} \rceil = 6\), \(e = 2(6) - 11.159 = 0.84\), and \(\Delta = \frac{0.84}{6} = 0.14\). Finally, \(y_t^{\text{adj}} = 4.312\), and the set \(DP\) is updated to \(DP \cup \{(1.443, 3.247), (3.303, 3.247), (5.163, 3.247), (7.022, 3.247), (8.882, 3.247), (10.742, 3.247)\}\).
In the third iteration, \(y_b\) is updated to \(y_t^{\text{adj}}\). Since \(y_b = 4.312\) is less than \(y_{\text{max}} = 8.5\), the AGD algorithm proceeds with the third iteration using \(y_b = 4.312\) and \(y_t = 6.312\). The set \(P\) is given by \(\{(1.014, 4.312), (11.725, 4.312), (1.485, 6.312), (10.499, 6.312), (12, 5)\}\), where \(P_1 = \{(1.014, 4.312), (11.725, 4.312)\}\), \(P_2 = \{(1.485, 6.312), (10.499, 6.312)\}\), and \(P_3 = \{(12, 5)\}\). Based on the points in set \(P\), we find \(x_{\text{min}} = 1.014\), \(x_{\text{max}} = 12\), and \(l = 10.985\). Therefore, \(n = 6\), \(e = 1.014\), \(\Delta = 0.169\), and \(y_t^{\text{adj}} = 6.468\). Additionally, the set \(DP\) is updated to \(DP \cup \{(1.930, 5.390), (3.761, 5.390), (5.591, 5.390), (7.422, 5.390), (9.253, 5.390), (11.084, 5.390)\}\).
In the final iteration, \(y_b\) is updated to \(6.468\). Since \(6.468 < 8.5\), the AGD algorithm proceeds with this iteration using \(y_b = 6.468\) and \(y_t = 8.468\). To find the set \(P\), we obtain \(P_1 = \{(10.321, 6.468), (1.522, 6.468)\}\), \(P_2 = \{(8.035, 8.468), (1.992, 8.468)\}\), and \(P_3 = \emptyset\). Therefore, \(P = \{(10.321, 6.468), (1.522, 6.468), (8.035, 8.468), (1.992, 8.468)\}\). From the points in set \(P\), we have \(x_{\text{min}} = 1.522\) and \(x_{\text{max}} = 10.321\), so the maximum length of the decomposition channel is \(l = 8.799\). Next, we calculate \(n = 5\), \(e = 1.2\), \(\Delta = 0.24\), and \(y_t^{\text{adj}} = 8.682\). The set \(DP\) is updated to \(DP \cup \{(2.401, 7.575), (4.161, 7.575), (5.921, 7.575), (7.681, 7.575), (9.441, 7.575)\}\). By updating \(y_b\) to \(y_t^{\text{adj}} = 8.682\) for the next iteration, we observe that \(y_b > 8.5\), which indicates that the AGD algorithm stops. Finally, the set \(DP\) is returned as the output of the AGD algorithm.

\begin{figure}
    \centering
    \includegraphics[width=0.9\linewidth]{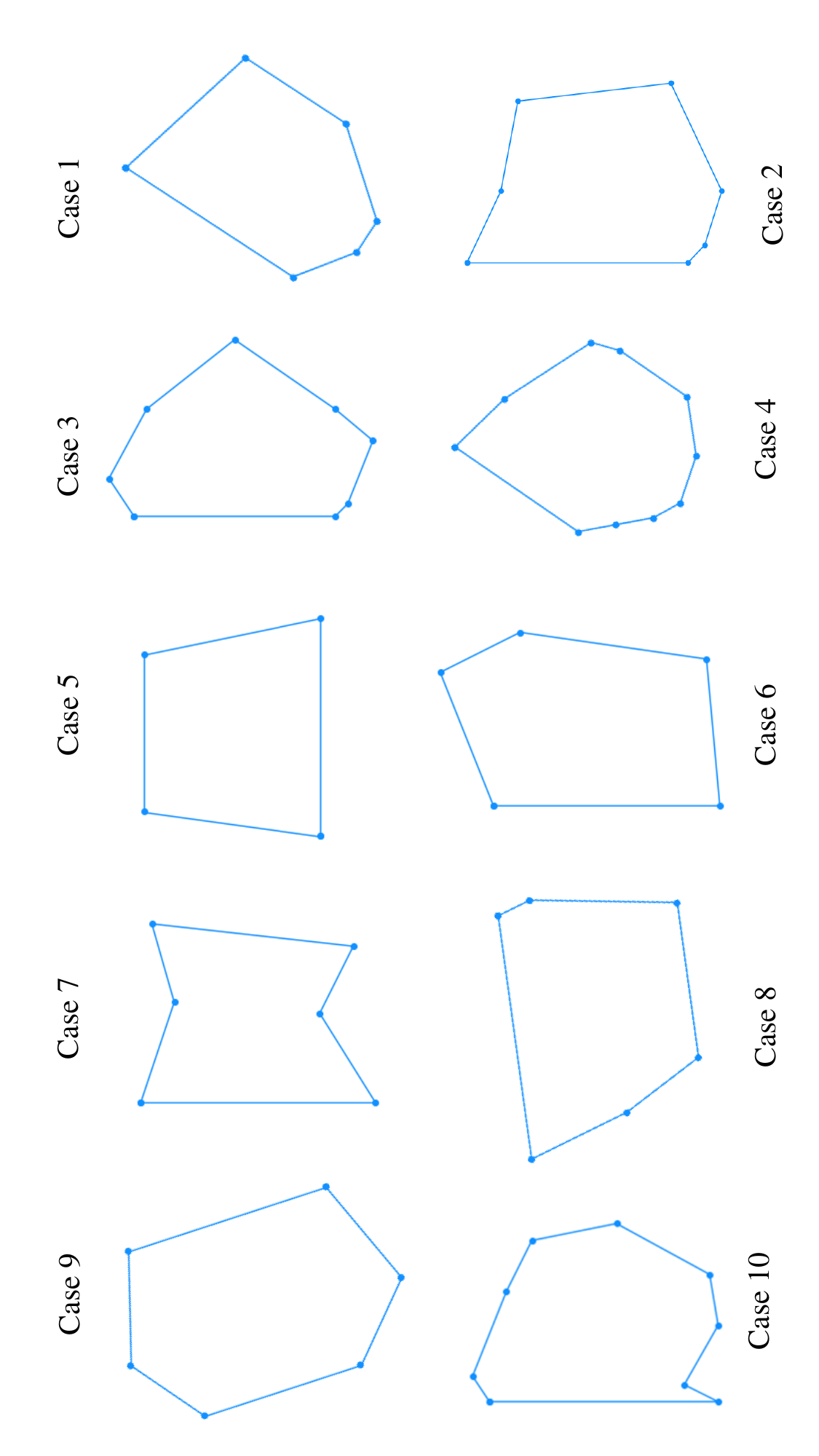}
    \caption{Cases in Table \ref{tab:case_result}.}
    \label{fig:Cases}
\end{figure}

The AGD algorithm can handle convex polygons. For non-convex polygons, the algorithm is first applied to their convex hull, and cells outside the shape are subsequently removed. However, the AGD algorithm can directly address some simple non-convex shapes, such as cases 2, 7, and 10 in Fig. \ref{fig:Cases}. Codes are available at \url{https://github.com/Sina14KD/AGD-CPP}.

\begin{table}
\caption{AGD algorithm notations} \label{tab:AGD notation} 
\begin{tabular}{@{}ll@{}}
\toprule
Notations & \\
\midrule
$r$ & Radius of UAV's camera footprint\\
$y_b$ &  Bottom line of the decomposition channel\\
$y_t$ & Top line of the decomposition channel\\
$y_t^{adj}$ & Adjusted top line of the decomposition channel\\
$l$ & Maximum length of the decomposition channel\\
$n$ & Number of cells in decomposition channel, i.e., $\lceil \frac{l}{\sqrt{2}r} \rceil$\\
$e$ & Excess length which are covered by $n$ cells, i.e., $\sqrt{2}rn-l$\\
$\Delta$ & Adjustment value, i.e., $\frac{e}{n}$\\
$DP$ & Set of cell center coordinates after decomposition\\

\botrule
\end{tabular}

\end{table}

\begin{algorithm}
\caption{AGD}\label{alg:phaseTwo}
\begin{algorithmic}[1]
\State \textbf{function} \ AGD (set of the polygon's vertices, r)
\State $y_b \gets 0$, $y_t \gets \sqrt{2}r$, $y_{max} \gets$ the maximum y-coordinate of the vertices, $DP \gets$ empty set

\While{$y_b < y_{max}$}  
     \State $P_1 \gets \text{set of points located at the intersection of } y_b \text{ and the polygon}$

    \State $P_2 \gets \text{set of points located at the intersection of } y_t \text{ and the polygon}$
    \State $P_3 \gets \text{set of polygon's vertices located between } y_b \text{ and } y_t$
    \State $P \gets P_1 \cup P_2 \cup P_3$
    \State $x_{min} \gets \text{the minimum x-coordinate of the points in } P$
    \State $x_{max} \gets \text{the maximum x-coordinate of the points in } P$
    \State $l \gets x_{max}-x_{min}$, $n \gets \lceil \frac{l}{\sqrt{2}r} \rceil$, $e \gets \sqrt{2}rn-l$
    \State $\Delta \gets \frac{e}{n}$
    \State $y^{adj}_t \gets y_b+\sqrt{2r^2+2\sqrt{2}r\Delta-\Delta^2}$
    \State $x^{\prime} \gets x_{min}+\frac{\sqrt{2}r-\Delta}{2}$
    \While{$x^{\prime}-\frac{\sqrt{2}r-\Delta}{2} < x_{max}$}

    \State $DP \gets DP \cup \{(x^{\prime}, y_b+\frac{y_t^{adj}-y_b}{2})\}$
    \State $x^{\prime} \gets x^{\prime}+\sqrt{2}r-\Delta$

    \EndWhile

    \State $y_b \gets y_t^{adj}$
    \State $y_t \gets y_t^{adj}+\sqrt{2}r$

\EndWhile
\State \textbf{return} $DP$

\end{algorithmic}
\end{algorithm}

\subsection{Mathematical model} \label{sect:MIP_model}
In this section, we propose a MIP model to address the CPP problem for a single UAV using the notations in Table \ref{tab:MIP notation}. For simplicity, the model determines the starting and final cells in the UAV's path and provides an estimation of the coverage time, though this estimation may not correspond to the best feasible path.

\begin{table}
\caption{MIP model notations} \label{tab:MIP notation} 
\begin{tabular}{@{}ll@{}}
\toprule
Notations & \\
\midrule
$I$ & Set of cells' indices in the set $DP$, i.e., $\{1,2,\dots,|DP|\}$.\\
$x_i$ & 1 if the UAV exits from the \(i^{\text{th}}\) cell; otherwise, 0.
\\
$z_i$ &  1 if the UAV enters the $i^{\text{th}}$ cell; otherwise 0.\\
$y_{ij}$ & 1 if the UAV travels directly from the \(i^{\text{th}}\) cell to the \(j^{\text{th}}\) cell without stopping in any other cells;\\ &otherwise, 0.\\
$d_{ij}$ & Distance between cell $i$ and $j$.\\
$v$ & Airspped of the UAV.\\
$t_{\text{cov}}$ & Coverage time (the time required to fully cover the area).\\

\botrule
\end{tabular}
\end{table}

\begin{align}
    &Min &&t_\text{cov} &&
    \label{eq:mip_obj}\\
    &s.t. && x_{|I|}+z_{1}\le 0 \ \label{eq:mip_c1}
\\
    & && \sum_{i \in I}y_{ij}=z_j  \ \ \ \forall j \in I \  \label{eq:mip_c2}\\
    & &&  \sum_{j \in I}y_{ij}=x_i  \ \ \ \forall i \in I \label{eq:mip_c3} \\
    & && \sum_{i\in I }x_i+z_i \ge 2|I|-2\label{eq:mip_c4} \\ 
     & &&  y_{ij}+y_{ji}\le 1 \ \ \ \forall i,j \in I  \label{eq:mip_c5}\\ 
    & && t_{\text{cov}}= \frac{1}{v} \sum_{i \in I}\sum_{j \in J} d_{ij}y_{ij} \label{eq:mip_c6}\\
    & &&z_{i},x_{i}, y_{ij} \in \{0,1\} \ \ \ \forall i,j \in I
    \label{eq:mip_c7}
\end{align}

The objective function (\ref{eq:mip_obj}) minimizes coverage time, and Constraint (\ref{eq:mip_c1}) ensures that the UAV cannot exit the \( |I|^{\text{th}} \) cell (i.e., the last cell in the UAV's path), nor can it enter the first cell (\(i=1\)), as this is the starting point. Constraint (\ref{eq:mip_c2}) ensures that the UAV cannot enter the \(i^{\text{th}}\) cell from more than one cell, and Constraint (\ref{eq:mip_c3}) guarantees that the UAV cannot exit the \(i^{\text{th}}\) cell to more than one cell. Constraint (\ref{eq:mip_c4}) ensures the connectivity of the UAV's path and that all cells are visited (full coverage). Constraint (\ref{eq:mip_c5}) prevents any loops between two cells. Finally, Constraint (\ref{eq:mip_c6}) calculates the coverage time for the UAV's path.

\begin{figure}
    \centering
    \includegraphics[width=0.9\linewidth]{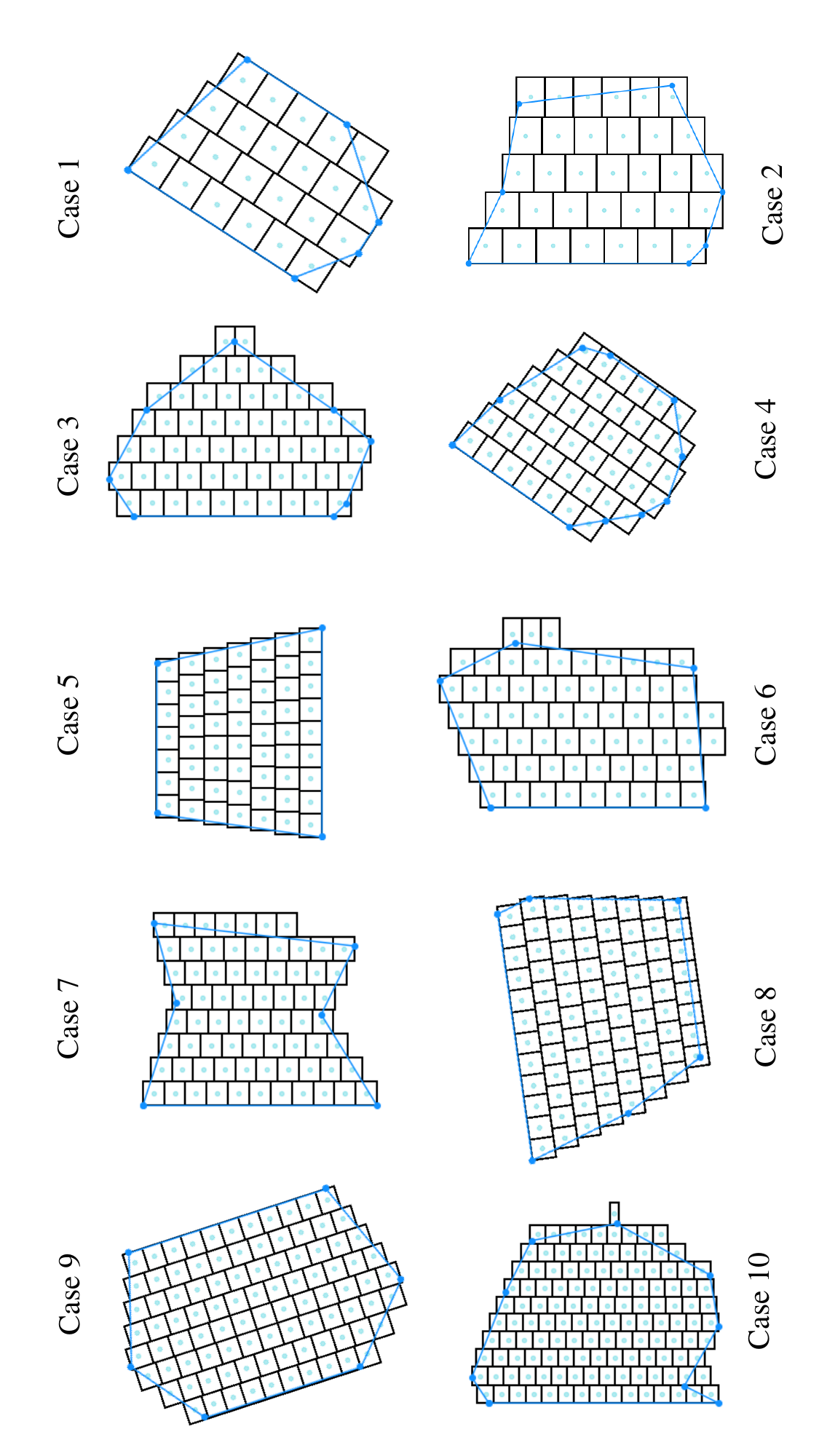}
    \caption{Cases in Table \ref{tab:case_result} after the AGD algorithm implementation.}
    \label{fig:Cases-after-AGD}
\end{figure}

\section{Experiments and Results} \label{sec:experimental}

In this section, we compare the performance of the Adaptive Grid-based Decomposition (AGD) algorithm with the Standard Grid-based Decomposition (SGD) method to demonstrate the efficiency of our proposed algorithm. First, we provide an overview of the SGD method, followed by a presentation of the comparison results.

In the Standard Grid-based Decomposition (SGD) method, to maintain consistency with the AGD algorithm process, we first transform the polygon so that its longest edge lies on the X-axis, and all vertices are positioned in the first quadrant of the XY-coordinate system (an example of the transformation process is detailed in Example 1 in the Appendix). Next, a grid with cells of edge length \(\sqrt{2}r\) (\(r\) being the radius of the camera footprint, see Fig. \ref{fig:footprint&grid}) is overlaid on the polygon. The bottom edge of the grid aligns with the X-axis, and the left edge aligns with the leftmost vertex of the polygon (the vertex with the minimum X-coordinate). We then retain only the cells that lie entirely or partially within the polygon and record their centers. To retrieve the original coordinates of these cell centers, the reverse transformation is applied. For instance, Figure \ref{fig:AGD-algorithm} illustrates the standard grid-based decomposition of a polygon with vertices \(\{(0, 0), (10, 0), (12, 5), (8, 8.5), (2, 8.5)\}\) and \(r = \sqrt{2}\).

For the experiments, we set the UAV's camera footprint (\(r\)) to \(50\sqrt{2}\) meters and the UAV's airspeed (\(v\)) to \(12 \, \text{m/s}\). Using this configuration, we analyze 10 polygon cases, with the corresponding results presented in Table \ref{tab:case_result}. The first column in Table \ref{tab:case_result} lists the case numbers corresponding to Fig. \ref{fig:Cases}, while the second column provides the area of each case in square meters (\(m^2\)). The third column shows the number of cells required to cover the search area using the SGD method (\(N_{\text{SGD}}\)), and the fourth column shows the number of cells required using the AGD algorithm (\(N_{\text{AGD}}\)). The fifth column reports the reduction in the number of cells achieved by the AGD algorithm compared to the SGD method, calculated as \(N_{\text{SGD}} - N_{\text{AGD}}\). The sixth column calculates the lower bound of the coverage time for the SGD method using the formula \(Z_{\text{SGD}} = (N_{\text{SGD}} - 1) \times \frac{\sqrt{2}r}{v}\), where \(v\) is the UAV's airspeed. The seventh column presents the coverage time for the AGD algorithm (\(Z_{\text{AGD}}\)), computed using the MIP model described in Sect. \ref{sect:MIP_model}. Both coverage times are reported in seconds. The final two columns of Table \ref{tab:case_result} display the relative and absolute improvements achieved by the AGD algorithm over the SGD method.

The results show that the AGD algorithm can reduce coverage time by up to 20\%, improving the chances of a successful SAR mission. In particular, the AGD algorithm can save up to about 2 minutes in coverage time, a crucial advantage in maritime SAR where every second can make a difference. Additionally, the AGD algorithm reduces the number of cells required to cover the search area by up to 12 cells compared to the SGD method, further demonstrating its enhanced efficiency and optimization.

\begin{table}[h]
\caption{Results for test cases} \label{tab:case_result} 
\begin{tabular*}{\textwidth}{@{\extracolsep\fill}lcccccccc}
\toprule%
Case&Area&$N_{\text{SGD}}$ &$N_{\text{AGD}}$&cell& \multicolumn{2}{@{}c@{}}{Coverage time (s)} &Relative & Absolute \\\cmidrule{6-7}
 no.&($m^2$)& & &reduction & $Z_{\text{SGD}}$ & $Z_{\text{AGD}}$ & Improvement ($\%$)& Gap (s) \\
\midrule
1&208,750&29&23&6&233.32&184.57&20.9&48.75\\
2&285,000&35&33&2&258&283.32&8.9&25.32\\
3&493,125&61&57&4&500&450.04&9.9&49.96\\
4&393,325&51&44&7&416.66&348.75&16.2&67.91\\
5&561,875&69&58&11&566.66&459.87&18.8&106.79\\
6&556,250&69&65&4&575&515.66&10.3&59.34\\
7&613,750&75&71&4&616.66&571.79&7.2&44.87\\
8&762,500&86&82&4&708.33&660.04&6.8&48.29\\
9&895,625&107&95&12&883.33&768.75&12.9&114.58\\
10&1,172,500&137&130&7&1133.33&1056.25&6.8&77.08\\
\botrule
\end{tabular*}
\footnotetext{Relative improvement is calculated as \(\frac{Z_\text{SGD} - Z_\text{AGD}}{Z_\text{SGD}}\), and the absolute gap is calculated as \(|Z_\text{SGD} - Z_\text{AGD}|\). The cases are shown in Fig. \ref{fig:Cases} before decomposition and in Fig. \ref{fig:Cases-after-AGD} after the AGD implementation.}
\end{table}

\section{Conclusions}\label{sec:conclusion}

Search and rescue teams use UAVs to enhance the efficiency of SAR operations, where time is a critical factor—particularly in maritime SAR, where even seconds can be vital to saving lives. Thus, developing an approach to reduce UAV coverage time can significantly improve SAR efficiency and increase the likelihood of a successful rescue. We propose a new decomposition approach, the Adaptive Grid-based Decomposition (AGD) method, which partitions the search area into a grid with fewer cells than standard grid-based decomposition. Additionally, we review an MIP model adapted to the AGD method for single-UAV path planning. Experimental results demonstrate the AGD algorithm’s effectiveness in creating shorter coverage paths, thereby reducing overall coverage time (by up to $ 20 \%$).

Several aspects of this work can be expanded. First, the AGD algorithm can be used for multi-UAV path planning, which would benefit SAR teams with a fleet of UAVs. Second, the AGD algorithm could be applied to path planning in other contexts, with a focus on minimizing battery consumption.

\section*{Appendix}
In this section, we provide an example to demonstrate how to transform the polygon’s vertices before applying the AGD algorithm. While this is a straightforward process, Example 1 includes a detailed explanation for added clarity.\\

\begin{figure}[h]
    \centering
    \includegraphics[width=1\linewidth]{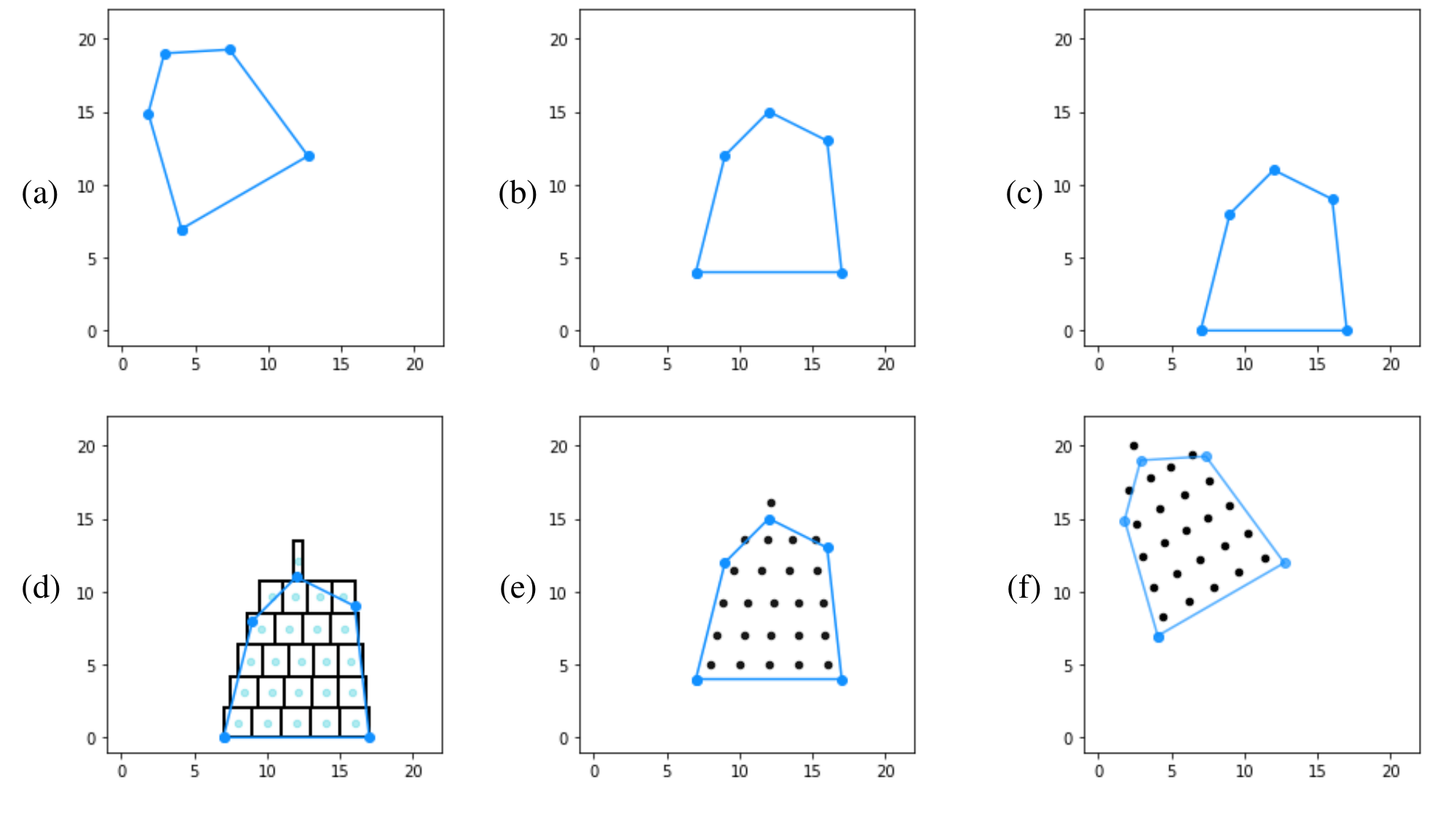}
  \caption{Example 1 in Appendix. (a) The main search area; (b) after applying the \( A \) matrix (30-degree clockwise rotation); (c) after applying the \( b \) vector (-4 shift in the Y-coordinate); (d) after applying the AGD algorithm; (e) after applying the \( -b \) vector (+4 shift in the Y-coordinate); (f) after applying the \( A^{-1} \) matrix (30-degree counterclockwise rotation to restore the main search area with the AGD decomposition).}

    \label{fig:Example_1}
\end{figure}

\textbf{Example 1.} 
Consider the set of points \(V_0=\{(4.06, 6.96), (12.72, 11.96), (7.35, 19.25), (2.89, 18.99),\\ (1.79, 14.89)\}\), which represent the vertices of a polygon defining the search area, as illustrated in Fig. \ref{fig:Example_1}(a). To decompose this area, we follow the steps below.

First, it is easy to see that rotating the polygon by 30 degrees clockwise will make its longest edge parallel to the X-axis. To achieve this, we use the transformation matrix $A$:
\[
A = 
\begin{pmatrix}
\cos(30^\circ) & \sin(30^\circ) \\ 
-\sin(30^\circ) & \cos(30^\circ)
\end{pmatrix}.
\]
Using the matrix $A$, the polygon is rotated, producing the shape shown in Fig. \ref{fig:Example_1}(b), with its vertices represented by the set \(V_1=\{x \in \mathbb{R}^2 \ |\  x^T=Ay^T \ \ \text{where}\ \ y\in V_0\}=\{(7, 4), (17, 4), (16, 13), (12, 15), (9, 12)\}\).
 Second, by applying the vector \(b = (0, -4)\), the polygon is shifted so that the longest edge aligns with the X-axis, and all vertices are positioned in the first quadrant of the XY-coordinate system, as shown in Fig. \ref{fig:Example_1}(c). The vertices of the shifted polygon are given by the set \(V_2=\{x \in \mathbb{R}^2 \ |\  x=y+b \ \ \text{where}\ \ y \in V_1\}=\{(7, 0), (17, 0), (16, 9), (12, 11), (9, 8)\}\).
Third, the AGD algorithm is applied to the polygon with \( V_2 \) vertices, dividing it into 24 cells, as illustrated in Fig. \ref{fig:Example_1}(d). The coordinates of the cell centers are stored in the \( DP \) set, as defined in Table \ref{tab:AGD notation}.
Fourth, we need to perform the reverse shift using \( -b = (0, 4) \), as seen in Fig. \ref{fig:Example_1}(e).
Finally, the reverse transformation is completed by rotating the polygon 30 degrees counterclockwise using the \( A^{-1} \) matrix:  
\[
A^{-1} = 
\begin{pmatrix}
\cos(30^\circ) & -\sin(30^\circ) \\ 
\sin(30^\circ) & \cos(30^\circ)
\end{pmatrix}
\]  
This yields the shape shown in Fig. \ref{fig:Example_1}(f).
At this stage, the AGD decomposition of the main search area in Fig. \ref{fig:Example_1}(a) is complete, and the centers of the cells are determined using:

\[
\begin{split}
&\{(4.43, 8.33),(6.16, 9.33),(7.89, 10.33),(9.62, 11.33),(11.36, 12.33),(3.76, 10.33),\\
& (5.37, 11.26),(6.98, 12.19),(8.59, 13.12),(10.19, 14.05), (3.06, 12.46),(4.53, 13.32),\\
&(6.01, 14.17),(7.48, 15.02),(8.96, 15.87),(2.56, 14.68),(4.23, 15.64),(5.89, 16.60),\\
&(7.56, 17.56),(2.10, 16.94),(3.53, 17.76),(4.96, 18.59), (6.38, 19.41),(2.42, 20.03)\}.\\  
\end{split}
\]

\printbibliography

\end{document}